\documentclass[runningheads]{llncs}
\usepackage{graphicx}

\begin{document}
\title{Data Centric Domain Adaptation for Historical Text with OCR Errors}
%
%

\author{Luisa M{\"a}rz\inst{1,2,4} \and
Stefan Schweter \inst{3} \and
Nina Poerner \inst{1}\and
Benjamin Roth \inst{2} \and
Hinrich Sch{\"u}tze \inst{1}
}
\authorrunning{März et al.}
%
\institute{Center for Information and Language Processing, Ludwig Maximilian University, Munich, Germany\\ \email{maerz@cis.lmu.de, inquiries@cislmu.org}\\
\url{https://www.cis.uni-muenchen.de/} \and
Digital Philology, Research Group Data Mining and Machine Learning,\\ University of Vienna, Austria
\and
Bayerische Staatsbibliothek M{\"u}nchen, Digital Library/Munich Digitization Center, Munich, Germany
\and
NLP Expert Center, Data:Lab, Volkswagen AG, Munich, Germany\\}
\maketitle              

\begin{abstract}
We propose new methods for in-domain and cross-domain Named
Entity Recognition (NER) on historical data for Dutch and French.
For the cross-domain case, we address domain shift by
integrating unsupervised in-domain data via contextualized
string embeddings; and OCR errors by injecting synthetic OCR
errors into the source domain and address data centric domain adaptation. We propose a general approach to imitate OCR errors in arbitrary input data. Our cross-domain as well as our in-domain results outperform
several strong baselines and establish state-of-the-art results. We publish preprocessed versions of
the French and Dutch Europeana NER corpora.

\keywords{Named Entity Recognition  \and Historical data \and \textsc{Flair}.}
\end{abstract}
\section{Introduction}
Neural networks achieve good NER accuracy on high-resource
domains such as modern news text or Twitter
\cite{akbik-etal-2019-pooled,baevski-etal-2019-cloze}. 
But on historical text, NER often performs poorly. 
This is due to several challenges:
i) Domain shift: Entities in historical texts can be different from contemporary entities, this makes it difficult for modern taggers to work with historical data.
ii) OCR errors: historical texts -- usually digitized by OCR -- contain systematic errors not found in non-OCR text \cite{DBLP:conf/icdar/Jean-CaurantTCB17}. In addition these errors can change the surface form of entities.
iii) Lack of annotation: Some historical text is now available in digitized form, but without labels, and methods are required for beneficial use of such data \cite{martinek}.

In this paper, we address \textit{data centric domain adaptation} for NER tagging on historical French and Dutch data. 
Following Ramponi and Plank \cite{ramponi2020neural}, data centric approaches do not adapt the model but the training data in order to improve generalization across domains.
We address both in-domain and cross-domain NER.
In the cross-domain setup, we use supervised contemporary data and integrate unsupervised historical data via contextualized embeddings.
We introduce artificial OCR errors into supervised modern data and find a way to perturb corpora in a general and robust way – independent of language or linguistic properties.

In the cross-domain setup as well as in-domain, our system outperforms neural and statistical state-of-the-art methods, achieving 69.3\% $F_1$ for French and 63.4\% for Dutch.
With the in-domain setup, we achieve 77.9\% for French and 84.2\% for Dutch. 
If we only consider named entities that contain OCR errors, our domain-adapted cross-domain tagger even performs better (83.5\% French/ 46.2\% Dutch) than in-domain training (77.1\% French/ 43.8\% Dutch).
Our main contributions are:
\begin{itemize}
    \item Release of the preprocessed French and Dutch NER corpora\footnote{\url{https://github.com/stefan-it/historic-domain-adaptation-icdar}};
    \item Developing synOCR to mimic historical data while exploiting the annotation of modern data;
    \item Training historical embeddings on a large amount of unlabeled historical data;
    \item Ensembling a NER system that establishes SOTA results for both languages and scenarios.
\end{itemize}

\section{Methods}
\subsection{Architecture} \label{architecture}
We use the \textsc{Flair} NLP framework \cite{akbik-etal-2019-flair}.
\textsc{Flair} taggers achieve SOTA results on various benchmarks and are well suited for NER. Secondly, there are powerful \textsc{Flair} embeddings. They are trained without explicit notion of words and model words as character sequences depending on their context. These two properties contribute to making atypical entities - even those with distorted surface - easier to recognize.
In all our experiments, word embeddings are generated by a character-level RNN and passed to a word-level bidirectional LSTM with a CRF as the final layer. Depending on the experiment, we concatenate ($\odot$) additional embeddings and refer to that as \textit{ensembling process}.

\subsection{Noise methods} \label{noise}
\begin{figure}
    \centering
        \caption{Example from the dutch train set. Text in its originial, the synOCR'd and the corrputed form.}
        \vspace*{0.5cm}
    \includegraphics[width=360pt]{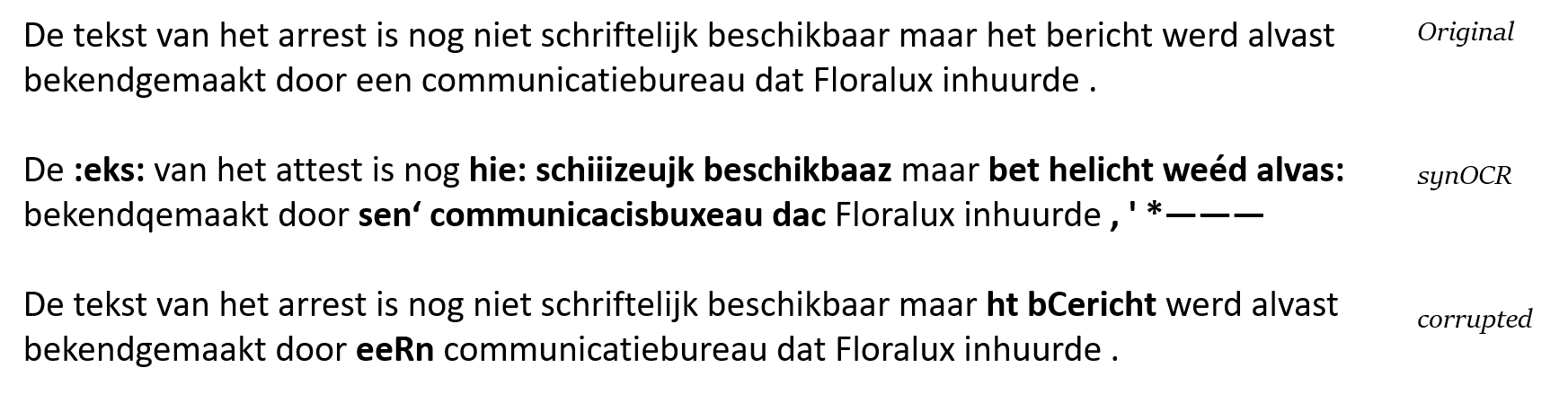}

    \label{example}
\end{figure}

Since digitizing by OCR introduces a lot of noise into the data, we recreate some of those phenomena in the modern corpora that we use for training. Our goal is to increase the similarity of historical (OCR'd) and modern (clean) data. An example drawn from the dutch training corpora can be found in Figure \ref{example}. Words that are different from the original text are indicated in bold font.\\

\textbf{Generation of synthetic OCR (synOCR) errors}
This method processes every sentence by assigning a randomly selected font and a font size between 6 and 11 pt.
Batches of 150 sentences are printed to PDF documents and then converted to PNG images.
The images are perturbed using
imgaug\footnote{\url{https://github.com/aleju/imgaug}} with the following steps:
(i) rotation,  
(ii) Gaussian blur and 
(iii) white or black pixel dropout.
The resulting image is recognized using tesseract version 0.2.6.\footnote{\url{https://github.com/tesseract-ocr/}}.
We re-align the recognized sentences with the clean annotated corpus to transfer the NER tags. For the alignment between original and degraded text we select a window of the bitext and calculate a character-based alignment cost. We then use the Wagner-Fisher algorithm \cite{fischer} to obtain the best alignment path through the window and the lowest possible cost. If the cost is below a threshold, we shift the window to the mid-point of the discovered path. Otherwise, we iteratively increase the window size and re-align, until the threshold criterion is met. This procedure allows us to find an alignment with reasonable time and space resources, without risking to lose the optimal path in low-quality areas.
Finally this results in an OCR-error enhanced annotated corpus with a range of recognition quality, from perfectly recognized to fully illegible. 
We refer to OCR-corrupted data as \emph{synOCR'd data}. \\

\textbf{Generation of synthetic corruptions}
This method is applied to our modern corpora, again to introduce noise as we find it in historical data.
Similar to \cite{DBLP:journals/corr/abs-1904-06707}, we
randomly corrupt 20\% of all words by 
(i) inserting a character or 
(ii) removing a character or 
(iii) transposing two characters. 
Therefore, we use the standard alphabet of French/ Dutch.
We re-align the corrupted tokens with the clean annotated tokens while maintaining the sentence boundaries to transfer the NER tags. Since the corruption method does not break the word boundaries we can simply map each corrupted word to the original one and retrieve the corresponding NER tag. 
We refer to synthetically corrupted data as \emph{corrupted data}.

\subsection{Embeddings}
\label{sec:embeddings}
We experiment with various common embeddings and integrate them in our neural system. Some of them are available in the community and some others we did train on data described in section \ref{data}.

\textbf{\textsc{Flair} embeddings}
\cite{akbik2018coling} present contextual
string embeddings which can be extracted from a neural language model. 
\textsc{Flair} embeddings use the internal states of a trained character language model at token boundaries. 
They are contextualized because a word can have different embeddings depending on its context.
These embeddings are also less sensitive to misspellings and rare words and can be learned on unlabeled corpora. 
We also use \textit{multilingual \textsc{Flair} embeddings}. 
They were trained on a mix of corpora from different domains (Web, Wikipedia, Subtitles, News) and languages.

\textbf{Historical embeddings}
We train \textsc{Flair} embeddings on large unlabeled historical corpora from a comparable time period (see Section \ref{unlabeled hist data}) and refer to them as \textit{historical embeddings}.

\textbf{BERT embeddings}
Since \textit{BERT embeddings} \cite{devlin-etal-2019-bert} produce state-of-the-art results for a wide range of NLP tasks, we also experiment with multilingual BERT embeddings\footnote{We use the cased variant from \url{https://huggingface.co/bert-base-multilingual-cased}}. 
BERT embeddings are subword embeddings based on a bidirectional transformer architecture and can model the context of a word. 
For NER on CoNLL-03 \cite{conll03}, BERT embeddings do not perform as well as on other tasks \cite{devlin-etal-2019-bert} and we want to examine if this observation holds for a cross-domain scenario with different data.

\textbf{FastText embeddings}
We do also use \textit{FastText embeddings} 
\cite{bojanowski-etal-2017-enriching} which are widely used
in NLP. They can be efficiently trained and address character-level phenomena. Subwords are used to represent the target word (as a sum of all its subword embeddings).
We use pre-trained FastText embeddings for French/Dutch\footnote{\url{https://fasttext.cc/docs/en/crawl-vectors.html}}.

\textbf{Character-level embeddings}
Due to the OCR errors out-of-vocabulary problems occur. Lample et al. \cite{lample-etal-2016-neural} create \textit{character embeddings}, passing all characters in a sentence to a bidirectional LSTM. 
To obtain word representations, the forward and backward representations of all the characters of the word from this LSTM are concatenated. Having the character embedding, every single words vector can be formed even if it is out-of-vocabulary. Therefore, we do also compute these embeddings for our experiments.

\section{Experiments}
In the cross-domain setup, we train on modern data (clean or synOCR'd) and test on historical data (OCR'd).
In the in-domain setup, we train and test on a set of historical data (OCR'd). 
We do use different combinations of embeddings and also use our noise methods in the experiments.

\subsection{Data} \label{data}
We use different data sources for our experiments from which some are openly available and some historical data come from an in-house project. For an overview of different properties (domain, labeling, size, language) see Table \ref{tab:data_props}.

\subsubsection{Annotated historical data}
Our annotated historical data comes from the Europeana Newspapers collection\footnote{\url{http://www.europeana-newspapers.eu/}}, which contains historical news articles in 12 languages published between 1618 and 1990.
Parts of the German, Dutch and French data were manually annotated with NER tags in IO/IOB format for PER (person), LOC (location), ORG (organization) by Neudecker \cite{neudecker-2016-open}.
Each NER corpus contains 100 scanned pages (with OCR accuracy over 80\%), amounting to 207K tokens for French and 182K tokens for Dutch.

We preprocess the data as follows.
We perform sentence splitting, filter out metadata, re-tokenize punctuation and convert all annotations to IOB1 format.
We split the data 80/10/10 into train/dev/test.
We will make this preprocessed version available in CoNLL format. 

\subsubsection{Annotated modern data}
For the French cross-domain experiments, we use the French WikiNER corpus \cite{WikiNER}.
WikiNER is tagged in IOB format with an additional MISC
(miscellaneous) category; we convert the tags to our Europeana format.
For better comparability we downsample (sentence-wise) the corpus from 3.5M to 525K tokens. Therefore, entire sentences were sampled uniformly at random without replacement.
For Dutch, we use the CoNLL-02 corpus
\cite{tjong-kim-sang-2002-introduction}, which consists of
four editions of the Belgian Dutch newspaper ``De Morgen'' from the year 2000.
The data comprises 309K tokens and is annotated for PER, ORG, LOC and MISC.
We convert the tags to our Europeana format.

\subsubsection{Unlabeled historical data} \label{unlabeled hist data}
For historical French, we use ``Le Temps'', a journal published between 1861 and 1942 (initially under a different name), a similar time period as the Europeana Newspapers. 
The corpus contains 977M tokens and is available from the National Library of France.\footnote{\url{https://www.bnf.fr/fr}}
For historical Dutch, we use data from an in-house OCR project.
The data is from the 19th century and it consists of 444M tokens.
We use the unlabeled historical data to pre-train historical embeddings (see Section \ref{sec:embeddings}).

\begin{table}[tp]
    \centering
    \caption{Number of tokens per dataset in our experiments.}
    \label{tab:data_props}
    \begin{tabular}{c|l|l|c|c}
         & \textbf{domain} & \textbf{data} & \textbf{labeled} & \textbf{size} \\ \hline
         French & historical & Europeana NER & + & 207K \\
         & modern & WikiNER & + & 525K \\ 
         & historical & ``Le Temps'' & - & 977M \\ \hline
         Dutch & historical & Europeana NER & + & 182K \\
         & modern & CoNLL-02 & + & 309K \\
         & historical & in-house OCR & - & 444M 
    \end{tabular}
\end{table}

\subsection{Baselines}
We experiment with three baselines. (i) The Java implementation\footnote{\url{https://nlp.stanford.edu/software/CRF-NER.html}} of the Stanford NER tagger \cite{finkel-etal-2005-incorporating}.
(ii) A version of Stanford NER published by Neudecker
\cite{neudecker-2016-open}\footnote{\url{https://lab.kb.nl/dataset/europeana-newspapers-ner}} that was trained on Europeana. In contrast to our system they trained theirs on the entire amount of the labeled Europeana corpora with 4-fold cross validation.
(iii) NN base. The neural network (see Section \ref{architecture}) with FastText, character and
multilingual \textsc{Flair} embeddings, as recommended in Akbik et al.
\cite{akbik-etal-2019-flair}.
For French, we also list the result reported by \c{C}avdar \cite{cavdar}. 
Since we do not have access to their implementation and could not confirm that their data splits conform to ours, we could not compute the combined $F_1$ score or test for significance.

\begin{table}[tp]
    \centering
    \small
    \caption{Results ($F_1$ scores on French/Dutch Europeana test set) of training on Europeana French/Dutch training set. \emph{Hist. Embs.} are historical embeddings. Scores marked with * are significantly lower than \emph{NN base $\odot$ hist. Es}.} 
    \label{hist-hist}
    \begin{tabular}{l|c|c|c|c}

         \textbf{French Models} & overall & PER & ORG & LOC\\ \hline 
         \c{C}avdar & & 0.68 & 0.37 & 0.68  \\
         Stanford NER tagger & 0.662* & 0.569* & 0.335* & 0.753*\\
         Stanford Neudecker & 0.750* & 0.750* & \textbf{0.505} & 0.826*\\ 
         NN base & 0.741* & 0.703* & 0.320* & 0.813*\\ 
         NN base $\odot$ hist. Embs. & \textbf{0.779} & \textbf{0.759} & 0.498 & \textbf{0.832}\\ \hline
    \end{tabular}
    	
		\hspace*{1 cm}
		\newline
    	
        \begin{tabular}{l|c|c|c|c}
         \textbf{Dutch Models} & overall & PER & ORG & LOC\\ \hline 
         Stanford NER tagger & 0.696* & 0.640* & 0.333* & 0.794*\\ 
         Stanford Neudecker & 0.623* & 0.700* & 0.253* & 0.702* \\ 
         NN base & 0.818* & 0.809* & 0.442* & 0.871*\\ 
         NN base $\odot$ hist. Embs. & \textbf{0.842} & \textbf{0.833} & \textbf{0.480} & \textbf{0.891}\\ \hline
    \end{tabular}
\end{table}

\section{Results and Discussion}
We evaluate our systems using the CoNLL-2000 evaluation script\footnote{\url{https://www.clips.uantwerpen.be/conll2000/chunking/conlleval.txt}}, with $F_1$ score.
To check statistical significance we use randomized testing \cite{yeh-2000-accurate} and results are considered significant if $p < 0.05$. 

\subsection{In-domain setup}
For both languages we achieve the best results with NN base $\odot$ historical embeddings.
With this setup we can produce $F_1$ scores of around 80\% for both languages, which outperforms all three baselines in the overall performance significantly. 
The results are presented in Table \ref{hist-hist}.
For French, the overall $F_1$ score as well as the $F_1$ for LOC
and ORG is best with NN base $\odot$ historical embeddings.
For  ORG the pre-trained tagger of Neudecker \cite{neudecker-2016-open} works best, which could be due to the gazetteer information they included and of course due to the fact that they train with the entire Europeana data. We hypothesize that the category with the most structural changes over time is ORG. In the military or ecclesiastical context in particular, there are a number of names that no longer exist (in this form).
For Dutch we observe the best overall performance with NN base $\odot$ historical embeddings except for all entity types.

\subsection{Cross-domain setup}
As shown in Table \ref{wiki-hist}, NN base performs better than the statistical Stanford NER baseline, which is in line with the observations for the in-domain training.
We experimented with concatenating BERT embeddings to NN base.
For both languages this increases the performance (Table \ref{wiki-hist}, NN base $\odot$ BERT). 
The usage of the historical embeddings is also very beneficial for both languages.
We can achieve our best results by using BERT for Dutch and by using historical embeddings for French. We conclude that the usage of modern pre-trained language models is crucial for the performance of NER taggers.

We generated synthetic corruptions for the WikiNER/CoNLL corpus. 
This could not outperform NN base for both languages. 
The training on synOCR'd WikiNER/CoNLL gives slightly worse results than NN base too. 
The corruption of the training data without the usage of any embeddings seems to harm performance drastically, what is in line with the observation of Hamdi et al. \cite{hamdi}.
It is striking that the training on corrupted/synOCR'd Dutch gives especially bad results for PER compared to French. A look at the Dutch test set shows that many entities are abbreviated first names (e.g. in \textit{\textbf{A J} van Roozendal}) and are often misrecognized what leads to a performance decrease.  
\begin{table}[tp]
    \centering
    \small
    \caption{Results of training on WikiNER/CoNLL corpus. Scores marked with * are significantly lower than \emph{NN ensemble}.}
    \label{wiki-hist}
    \begin{tabular}{l|c|c|c|c}
         \textbf{French Models} & overall & PER & ORG & LOC\\ \hline
         \c{C}avdar & & 0.48 & 0.11 & 0.56 \\
         Stanford NER tagger & 0.536* & 0.451* & 0.059* & 0.618*\\
         NN base & 0.646* & 0.636* & 0.096* & 0.721*\\ \hline
         NN base $\odot$ BERT & 0.660 & 0.639* & \textbf{0.163} & 0.725* \\
         NN base $\odot$ hist.Embs. & 0.672 & 0.661* & 0.015* & 0.748\\
         corrupted WikiNER & 0.627* & 0.635* & 0.085* & 0.710*\\
         synOCR'd WikiNER & 0.619* & 0.590* & 0.078 & 0.710\\
         NN ensemble corrupted & \textbf{0.693} & 0.624 & 0.063 & \textbf{0.783} \\
         NN ensemble synOCR & 0.684 & \textbf{0.710} & 0.111 & 0.744\\
    \end{tabular}
    
		\hspace*{1 cm}
		\newline

	   \begin{tabular}{l|c|c|c|c}
         \textbf{Dutch Models} & overall & PER & ORG & LOC\\ \hline
         Stanford NER tagger & 0.371* & 0.217* & 0.083* & 0.564*\\
         NN base & 0.567* & 0.493* & 0.085* & 0.700*\\ \hline
         NN base $\odot$ BERT & \textbf{0.634} & \textbf{0.572} & \textbf{0.250} & 0.771 \\
         NN base $\odot$ hist. Embs. & 0.632 & 0.568 & 0.084 & 0.738*\\
         corrupted CoNLL & 0.535* & 0.376* & 0.155* & 0.717*\\
         synOCR'd CoNLL & 0.521* & 0.327* & 0.061* & 0.721*\\
         NN ensemble corrupted & 0.606* & 0.439* & 0.158 & \textbf{0.799} \\
         NN ensemble synOCR & 0.614 & 0.481 & 0.157 & 0.775\\
    \end{tabular}

\end{table}
For French the combination of NN base and historical embeddings, trained on corrupted data or on synOCR'd (\textit{NN ensemble corrupted/ NN ensemble synOCR'd}) gives the best results and outperforms all other systems. 
For Dutch \textit{NN ensemble corrupted} and \textit{NN ensemble synOCR} give slightly worse results than NN base $\odot$ BERT and NN base $\odot$ historical embeddings, but performs better than the tagger trained on synOCR'd or corrupted data only (Table \ref{wiki-hist}, NN ensemble).

\subsubsection{Ablation study}
We analyze our results and examine the composition of NN ensemble synOCR more closely (since the results for NN ensemble corrupted are very similar we perform the analysis for NN ensemble synOCR as a representative for both NN ensemble). 

The ablation study (see Table \ref{ablation}) shows that NN ensemble benefits from different information in combination. 
For French NN ensemble gives the best results only for PER. 
The overall performance increases if we do not use character level embeddings.
There is a big performance loss if we omit the historical embeddings. 
If we do not train on synOCR'd data the performance decreases.
\begin{table}[tp]
    \centering
    \small
    \caption{Ablation study. Results of training on the clean and the synOCR'd WikiNER/CoNLL corpus.}
    \label{ablation}
    \begin{tabular}{l|c|c|c|c}
         \textbf{French Models} & overall & PER & ORG & LOC\\ \hline
         NN ensemble synOCR & 0.684 & \textbf{0.710} & 0.011 & 0.744\\
         - char & \textbf{0.693} & 0.681 & 0.078 & \textbf{0.758}\\ 
         - word & 0.686 & 0.664 & 0.080 & 0.756\\
         - hist. Embs. & 0.619 & 0.590 & 0.078 & 0.710\\
         - synOCR'd data & 0.672 & 0.661 & \textbf{0.015} & 0.748\\
    \end{tabular}
    
        \hspace*{1 cm}
		\newline
	
	  \begin{tabular}{l|c|c|c|c}
         \textbf{Dutch Models} & overall & PER & ORG & LOC\\ \hline
         NN ensemble synOCR & 0.614 & 0.382 & \textbf{0.157} & 0.775\\
         - char & 0.600 & 0.404 & 0.119 & \textbf{0.780} \\ 
         - word & 0.584 & 0.430 & 0.102 & 0.745\\
         - hist. Embs. & 0.521 & 0.327 & 0.061 & 0.721\\
         - synOCR'd data & \textbf{0.632} & \textbf{0.568} & 0.084 & 0.738\\
    \end{tabular}
\end{table}
For Dutch we can observe these facts even more clearly. If we do not train on synOCR'd data the $F_1$ score even increases. 
If omitting the historical embeddings we loose performance as well.

To find out why our implementation of the assumption -- synOCR increases the similarity of the data and improves results -- does not have the expected effect, we analyze the test sets.
It shows, that only 10\% of the French and 6\% of the Dutch entities contain OCR errors.
Therefore the wrong predictions are mostly not due to the OCR errors, but due to the inherent difficulty of recognizing entities cross-domain. 
This also explains why synthetic noisyfication does not consistently improve the system.
In addition there are some illegible lines in the synOCR'd corpora consisting of dashes and metasymbols, what is not similar to real OCR errors. 

\begin{figure}
    \centering
    \caption{Example sentence from the French test set.}
    \vspace*{0.5cm}
    \includegraphics[width=350pt]{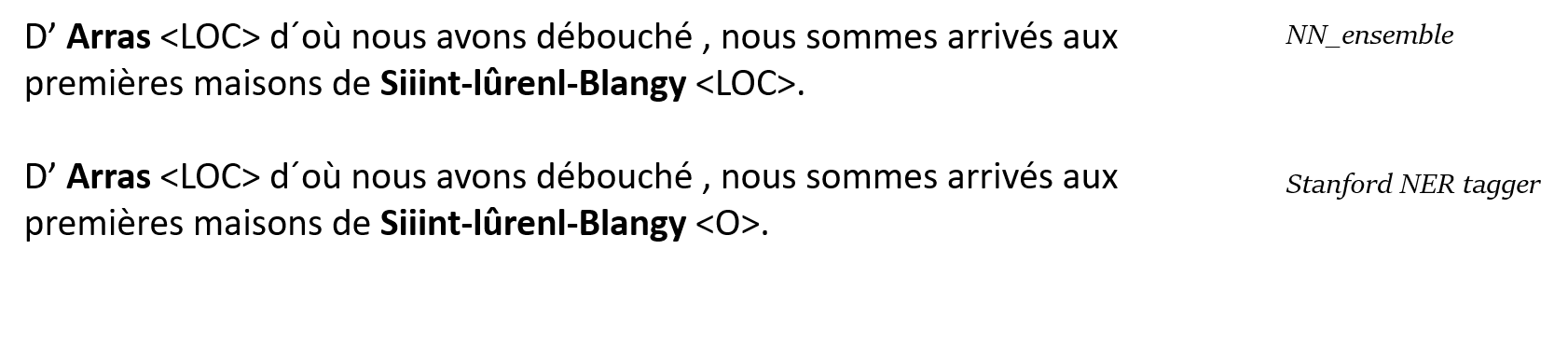}
    \label{example_2}
\end{figure}

To verify our assumption we also compare the different systems only on the entities with OCR errors. Here NN ensemble outperforms both of the cross-domain baselines (Table \ref{ocr}, Stanford NER tagger, NN base cross-domain).  Compared to the French results Dutch is a lot worse. A look at the entities shows that in the Dutch test set there are many hyphenated words where both word parts are labeled. However, if looking at the parts of the word individually, a clear assignment to an entity type cannot be made, which leads to difficulties with tagging.
Though it is plausible that NN ensemble can capture specific phenomena in the historical data better, since the difference between the domains is reduced by the synthetic noisyfication and the historical embeddings. 
The example in Figure \ref{example_2} drawn from the test set shows, that NN ensemble can handle noisy entities well in contrast to e.g. the Stanford NER tagger. 
Thus in a scenario with many OCR errors the NN ensemble performs well.

\begin{table}[tp]
    \centering
    \small
    \caption{Results on entities with OCR errors in the French/Dutch test set. Scores marked with * are significantly lower than \emph{NN ensemble}.}
    \label{ocr}
    \begin{tabular}{l|c|c}
         \textbf{Models} & French & Dutch\\ \hline
         Stanford NER tagger & 0.661* & 0.207*  \\
         NN base in-domain & 0.771 & 0.438 \\
         NN base cross-domain & 0.783 & 0.200* \\
         NN ensemble synOCR & \textbf{0.835} & \textbf{0.462} \\ 
    \end{tabular}
\end{table}

\section{Related Work}
There is some research on using natural language processing for improving OCR for historical documents \cite{berg-kirkpatrick-etal-2013-unsupervised,Vobl:2014:POS:2595188.2595197} and also on NER for historical documents \cite{ehrmann_2020_4117566}. In the latter - a shared task for Named Entity Processing in historical documents - Ehrmann et al. find that OCR noise drastically harms systems performance. Like us several participants (e.g. \cite{boros:hal-03026969}, \cite{DBLP:conf/clef/SchweterM20}) also use language models that were trained on historical data to boost the performance of NER taggers. 
Schweter and Baiter \cite{schweter-baiter-2019-towards} explore NER for historical German data in a cross-domain setting.
Like us, they train a language model on unannotated in-domain data and integrate it into a NER tagger.
In addition to the above mentioned work, we employ ``OCR noisyfication'' (Section \ref{noise}) and examine the influence of different pretrained embeddings systematically.
\c{C}avdar \cite{cavdar} addresses NER and relation extraction on the French Europeana Newspaper corpus. 
Ehrmann et al. \cite{Ehrmann2016DiachronicEO} investigate the performance of NER systems on Swiss historical Newspapers and show that historical texts are a great challenge compared to contemporary texts. They find that the LOC class entities causes the most difficulties in the recognition of named entities. 
The recent work of Hamdi et al. \cite{hamdi} investigates the impact of OCR errors on NER. To do so, they also perturb modern corpora synthetically with different degrees of error rates. They experiment with Spanish, Dutch and English. Like us they perturb the Dutch CoNLL corpus and train NER taggers on that data. Unlike us they do also train on a subset of the perturbed corpus. We test on a subset of the Dutch Europeana corpus. Hamdi et al. \cite{hamdi} show that neural taggers perform better compared to other taggers like the Stanford NER tagger and they also prove that performance decreases drastically if the OCR error rate increases. 
Piktus et al. \cite{piktus-etal-2019-misspelling} learn misspelling-oblivious FastText embeddings from synthetic misspellings generated by an error model for part-of-speech tagging.
We use a similar corruption method, but we also use synOCR and historical embeddings for NER.

\section{Conclusion}
We proposed new methods for in-domain and cross-domain Named Entity Recognition (NER) on historical data and addressed data centric domain adaptation.
For the cross-domain case, we handle domain shift by integrating non-annotated
historical data via contextualized string embeddings; and OCR errors by injecting synthetic OCR errors into the modern data. 
This allowed us to get good results when labeled historical data is not available and the historical data is noisy. For training on contemporary corpora and testing on historical corpora we achieve new
state-of-the-art results of 69.3\% on French and 63.4\% on Dutch.
For the in-domain case we obtain state-of-the-art results of 77.9\% for French and 84.2\% for Dutch.
There is an increasing demand for advancing the digitization of the world's cultural heritage.
High quality digitized historical data, with reliable meta information, will facilitate convenient access and search capabilities, and allow for extensive analysis, for example of historical linguistic or social phenomena.
Since named entity recognition is one of the most fundamental labeling tasks, it would be desirable that advances in this area translate to other labeling tasks in processing of historical data as well. 

\section{Acknowledgement}
This work was funded by the European Research Council (ERC \#740516).

%
%
\bibliographystyle{splncs04}
\bibliography{mybibliography}

\newpage
\appendix

\section{Appendix}
\label{sec:appendix}
\textbf{Detailed information about experiments and data}

\noindent The computing infrastructure we use for all our experiments is one GeForce GTX 1080Ti GPU with an average runtime of 12 hours per experiment.
For the French and Dutch baseline model \textit{NN base} we count 15,895,683 parameters each. For the French \textit{NN ensemble} model there are 88,264,777 parameters and 96,895,161 parameters for the Dutch \textit{NN ensemble}.\\ 
The Europeana Newspaper Corpus is split 80/10/10 into train/dev/test.
The downsampled French WikiNER corpus is split 70/15/15 into train/dev/test and the Dutch CoNLL-02 corpus is already split in its original version. 
A link to the downloadable version of the French and Dutch Europeana data will be included in the camera-ready version of the paper. 

\begin{table}
    \centering
    \small
    \caption{Number of tokens for each datasplit.}
    \label{stat2}
    \begin{tabular}{l|c|c|c}
         \textbf{Dataset} & train & dev & test\\ \hline
         French Europeana & 167,723 & 18,841 & 20,346  \\
         Dutch Europeana & 147,822 & 16,391 & 18,218\\
         French WikiNER & 411,687 & 88,410 & 88,509\\
         Dutch ConNLL-02 & 202,930 & 68,994 & 37,761 \\ 
    \end{tabular}	
\end{table}

\end{document}